\documentclass[conference]{IEEEtran}

\usepackage{graphicx,flushend,color}

\usepackage{amsmath,amsthm}
\usepackage{amssymb}

\usepackage[ruled,vlined]{algorithm2e}
\usepackage{soul}

\usepackage{fancyhdr}
\begin{document}
\title{Subtractive Perceptrons for Learning Images: \\A Preliminary Report}

\author{\IEEEauthorblockN{H.R.Tizhoosh$^{1,2}$, Shivam Kalra$^1$, Shalev Lifshitz$^1$,   Morteza Babaie$^1$}\\
$^1$ Kimia Lab, University of Waterloo,\\
$^2$ Vector Institute, Toronto, Canada
Ontario, Canada}

\maketitle

\begin{abstract}
In recent years, artificial neural networks have achieved tremendous success for many vision-based tasks. However, this success remains within the paradigm of \emph{weak AI} where networks, among others, are specialized for just one given task. The path toward \emph{strong AI}, or Artificial General Intelligence, remains rather obscure. One factor, however, is clear, namely that the feed-forward structure of current networks is not a realistic abstraction of the human brain. In this preliminary work, some ideas are proposed to define a \textit{subtractive Perceptron} (s-Perceptron), a graph-based neural network that delivers a more compact topology to learn one specific task. 
In this preliminary study, we test the s-Perceptron with the MNIST dataset, a commonly used image archive for digit recognition. The proposed network achieves excellent results compared to the benchmark networks that rely on more complex topologies.
\end{abstract}

\section{Motivation}
The quest for artificial intelligence has been focusing on the creation of an ultimate machine that can perform cognitive tasks at the level of human capabilities. In the past seven decades, tremendous effort has been focused on developing learning automatons that can learn from the past experience in order to generalize to unseen observations. After decades of slow progress and stagnation, we are finally witnessing the success of artificial neural networks (ANNs) mainly achieved though the concept of \emph{deep learning}. A large number of cognitive tasks have been performed with deep architectures to accurately recognize faces, objects and scenes. 

\begin{figure}[htb]
\begin{center}
\includegraphics[width=0.45\textwidth]{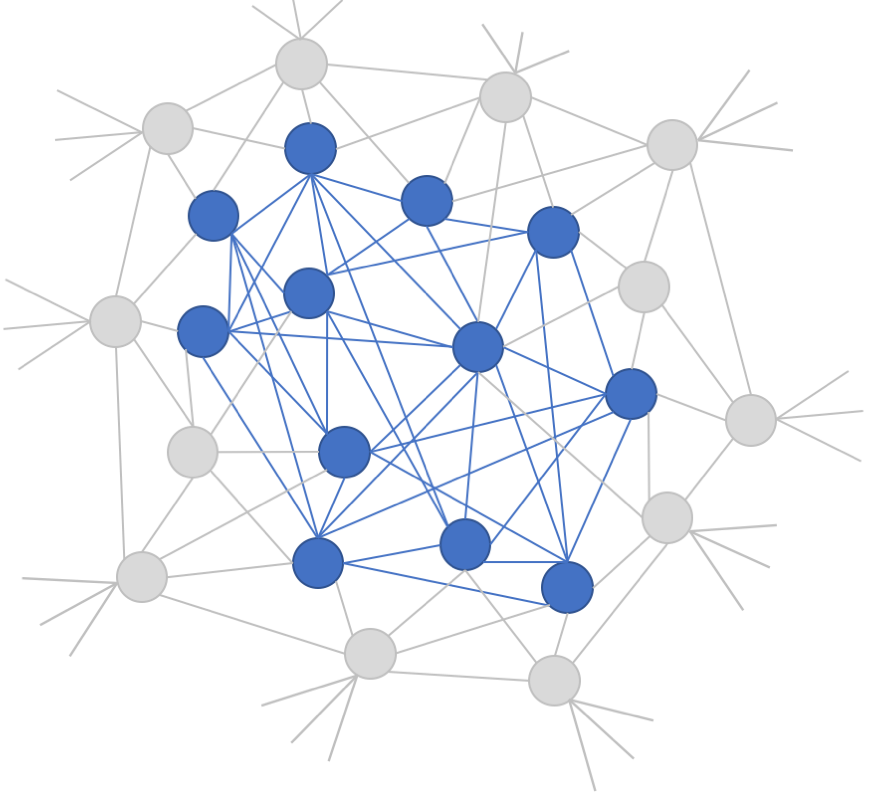}
\caption{Motivation: If we define the Perceptron as a graph (blue nodes; to perform one specific task) it may be more easily integrated within a larger graph (gray nodes; responsible for many other tasks). }
\label{fig:brain}
\end{center}
\end{figure}

The recent success has also made it clear  that the intrinsic nature of current ANNs is based on uni-task orientation \cite{tizhoosh2018artificial}. Besides, multiple ANNs cannot be easily \emph{integrated} into a larger network in order to perform multiple tasks as is probably the case in human brains (see Fig. \ref{fig:brain}) \cite{saaty2019brain,braun2015dynamic}. It is a well-known fact that the \emph{feed-forward} and \emph{layered} topology of ANNs is not a realistic model of how the human brain is structured. This has been the case since the early days of artificial intelligence when \emph{Perceptrons} were first introduced. Hence, revisiting the Perceptron as a building block of ANNs appears to be the right focus of any investigation to move toward more capable ANNs.    
In this paper, we introduce the notion of a \textit{subtractive Perceptron} (s-Perceptron), a graph-based neural network that uses shared edges between neurons to process input differences in addition to the ordinary inputs. The differences should not only provide more information on the input but also provide the values for additional weights, namely for the edges between the neurons. We test the proposed architecture with the MNIST dataset for digit recognition and critically summarize the findings. 

\section{Background}
ANNs have been under investigations for more than half a century. A fundamental limitation of ANNs is that they are a collection of feed-forward layers of neurons which is different from the rather graph-like and massively connected topology of human brain \cite{saaty2019brain}. This layered structure is implemented to simplify the learning for gradient descent and its variations to solve the credit assignment problem. However, ANNs can hardly claim to be a realistic imitation of the human brain. The latter is a massively connected network of neurons, approximately $80\times 10^{9}$ neurons with more or less $10^{14}$ connections, with an average of 10,000 connections per neuron. We are all aware of this ``model flaw'' but there is  not much we can do about it, and considering the recent success stories of deep ANNs we may not even have to. In addition, there have been other  compelling reasons to stick with feed-forward structures and not investigate graphs as a potential model extension. The most apparent reason is perhaps that our training graphs with neurons with mutual dependencies may be very difficult. Even breakthroughs such as layerwise pre-training are based on ``restricted'' Boltzmann machines, architectures that eliminate intra-layer connections to convert a general Boltzmann machine into a feed-forward network. 

The history of ANNs starts with the introduction of Perceptrons in the 1950s \cite{Ros58,Min17} which can be defined as  linear classifiers. The class of feed-forward multi-layer networks found attraction after several authors introduced the backpropagation algorithm was conceived and further developed \cite{bryson1969applied,Rum85}. Continuing up to the late 1990s, multi-layer Perceptrons (MLPs) were intensively investigated with many applications in classification, clustering and regression \cite{Zha00}. However, it was becoming increasingly clear that MLPs were providing low recognition rates for challenging problems when configured with a small number of layers, in which the increasing the number of layers would result in intractable training sessions, unable to deliver results \cite{Glo10}. Alternative architectures such as the Neocognitron \cite{Fuk88} with the goal of finding a solution but no capable learning algorithm was available to exploit a new topology. 

On the other hand, self-organizing maps were among the first generation of practical ANNs which could cluster data in an unsupervised manner \cite{Koh95}.
The introduction of convolutional neural networks (CNNs), was a major milestone toward a breakthrough \cite{Lec95}. CNNs used multi-resolutional representations and, through weight sharing, incorporated the learning of a multitude of filters to extract unique features of input images \cite{Ben13}. This ultimately solved the issue with external and non-discriminating features yet the training of CNNs remained a challenge. In subsequent years, several ideas changed this: the introduction of restricted Boltzmann machines, and the layer-wise greedy training enabled researchers to train ``deep networks''; CNNs with many hidden layers \cite{Hin06,Ben07}. The first impressive results of such networks for image recognition started the chain of success stories leading to what we call \emph{deep learning}  \cite{Kri12}. Of course, there still may be cases when a capable non-neural classifier combined with cleverly designed features might be the solution \cite{camlica2015medical} but the deep networks have drastically restricted the applications of such techniques.   

There currently exists a rather large body of literature on ``complex brain networks'' \cite{Spo04a,Spo04b,Boc06,Buc13}. The general idea of using ``graphs'' in connection with neural networks is not new. Kitano used graph notation to generate feed-forward networks and learn via evolutionary algorithms \cite{Kit90}. However, this had no effect on the topology of the network. Generalized random connectivity has been explored to shed light on more complex networks \cite{Str01}. Modelling the human brain as a collection of interacting networks has recently drawn attention in research as well \cite{Rei07,Sta07}. This may clarify anatomical and functional connectivity in the brain, and subsequently impact the way that we understand and design ANNs. In such networks that can be implemented as graphs, edges could be both weighted or unweighted \cite{Rei07}. 

The theories of ``small-world and scale-free networks'' can also contribute to the development of neuroscience, pathology and to the field of AI \cite{Sta07}. Small-world networks lay between regular and random graphs and are structurally close to social networks in that they have a higher clustering and almost the same average path than random networks with the same number of nodes and edges \cite{simard2005fastest}. In contrast  scale-free networks are heterogeneous with respect to degree distribution. They are called scale-free since when zooming in on any par, one can observe a few but significant number of nodes with many connections, and there is a trailing tail of nodes with very few connections at each level of magnification \cite{Nod18,Wan03}. So far, few papers have suggested to put this type of brain modelling into any implementation. The main challenge was, and still is, to propose a new  ``learning'' procedure for such complex networks. 

In 2009, Scarselli et al. suggested Graph Neural Networks (GNNs) \cite{Sca09}. The authors had applications such as computer vision and molecular biology in mind, but GNNs do in fact use specific algorithm for training, which is based on backpropagation like traditional architectures. This is because their proposed architecture is still of ``feed-forward'' nature and the ``graph'' aspect of their model was the capability of extending to the inputs available in graph forms, and not the internal topology of the network itself. Similarly, Duvenaud et al. apply deep networks on graphs to learn molecular fingerprints \cite{Duv15}. As well, Niepert et al. use neighbourhood graph construction and subsequent graph normalization to feed data into a convolutional network \cite{Nie16}. Multi-layer Graph Convolutional Network (GCN) also use graphs but they keep the general feed-forward architecture \cite{Kip17}. In contrast, Bruna et al. have recognized that incorporating ``manifold-like graphs'' structure can create more compact topologies (less parameters to adjust) \cite{Bru14}. They propose an exploitation of the graph's global structure  with the spectrum of its graph-Laplacian to generalize the convolution operator when convolutional networks are the base for the design. Hence, their graph understanding is rather implicit and restricted to grids in order to sustain the useful, but limiting, convolutional structure. Qi et al. demonstrated some usability of these networks for image segmentation \cite{Qi17} while Li et al. proposed the gated recurrent version of graph neural networks to process non-sequential data \cite{Li17}. 

In summary, although most scientists suggest that the brain is composed of densely connected (graph-like) structures \cite{braun2015dynamic,hosseini2012gat},  research has primarily focused on ANNs with a series of stacked layers such that the main ANN learning algorithm, gradient descent, can be applied through error backpropagation. Any innovation in the ANN domain, thus, requires us not only to fully embrace ``connected graphs'' as the building block for not-layered networks, but also to be prepared to abandon gradient descent and explore unknown territories in search of new learning algorithms that can handle the neuron dependencies as a result of intra-layer connections.

A network that similar to brain topology has to be a graph with a large number of connections. In the  section, the concept of a \emph{subtractive Perceptron} is introduced. Inputs are projected into a graph and the output of neurons are projected into output nodes. We will discuss the design challenges and attempt to provide solutions. Additionally, we run preliminary experiments to demonstrate the feasibility of our ideas.  

\section{Subtractive Perceptron}
\begin{figure}[htb]
\begin{center}
\includegraphics[width=0.9\columnwidth]{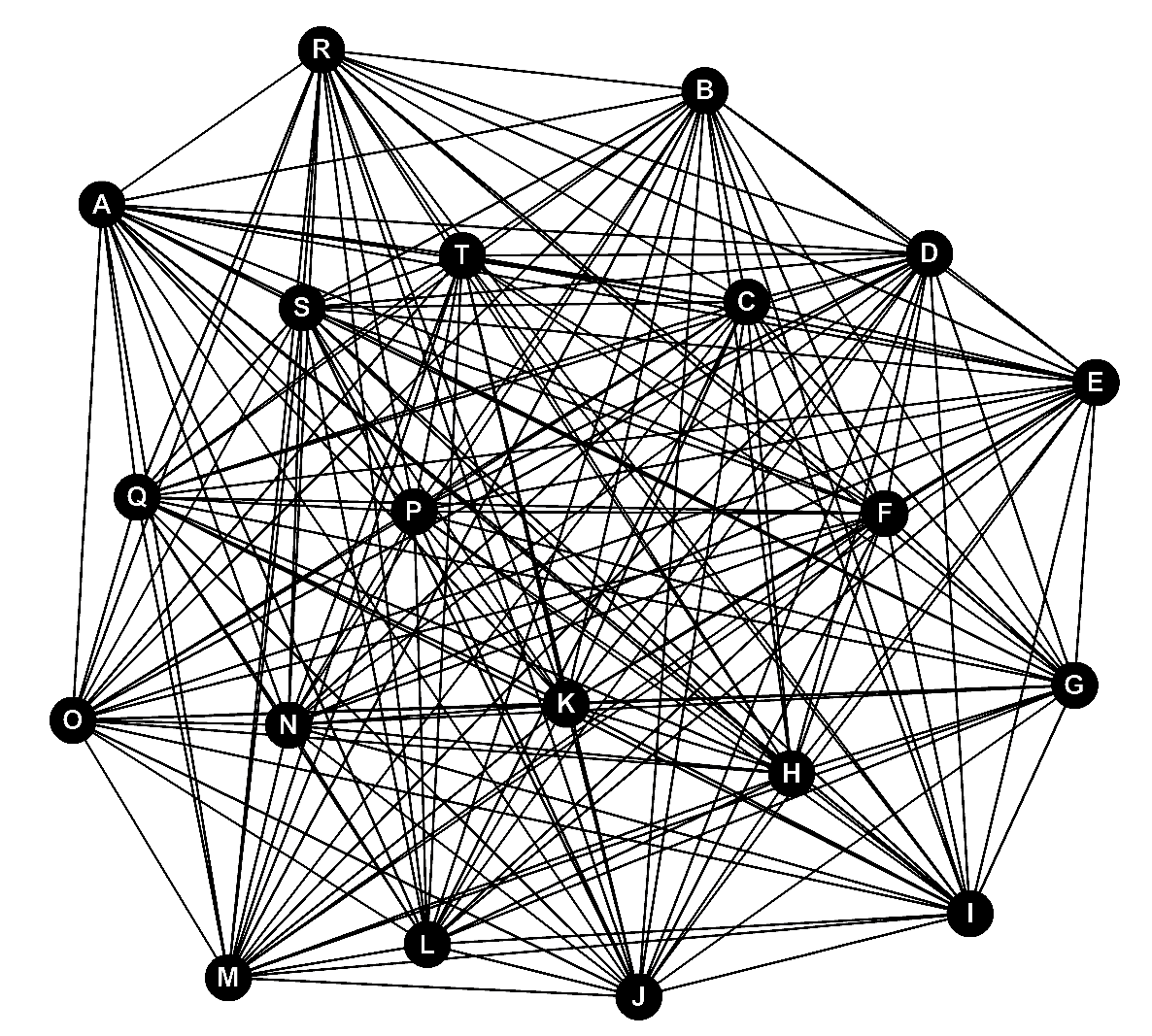}
\caption{A complete undirected graph $G=(V,E)$ with $n=|V|=20$ vertices (neurons/nodes) and $|E|=n(n-1)/2=190$ edges. All edges are assigned a weight that is shared between corresponding nodes. If  outputs (axons) of neurons are the inputs (dendrites) to other neurons, learning algorithms such as gradient descent may not be able to adjust the weights due to mutual dependencies.}
\label{fig:comgraph}
\end{center}
\end{figure}

Consider an undirected complete graph $G=(V,E)$ with a set of nodes $V$ and their edges $E$. This means for every randomly selected $v_i$ and $v_j$ ($i \neq j$), the edges $E(v_i,v_j)$ and $E(v_j,v_i)$ exist and share the same weight (Fig. \ref{fig:comgraph}). Such a network can be considered a small sub-graph within a much larger graph (i.e., similar to the brain, see Fig. \ref{fig:brain})  that each learn a specific task.  

The transition from a traditional Perceptron to its graph version  creates several design and algorithmic  problems that need to be addressed in order to create a new type of neural network. In the following sub-section, we explore these challenges and attempt to propose preliminary solutions. 

\subsection{Problem 1: In- and Outputs}
Working on a complete Graph $G$ with $n$ neurons, the first question would be how to organize the in- and outputs. As there are no layers, there is also no obvious access points for inputting data and outputting the results, something that are obvious and simple in feed-forward layered networks.

 Assuming that for $n$ neurons we have $n_\textrm{input}$ input neurons and $n_{output}$ output neurons, the most straightforward solution would be $n=n_\textrm{input}=n_{output}$.
 If we project all inputs into all neurons and we take an output from each neuron, this question can be easily answered. Taking as many neurons as we have inputs and outputs may be convenient but will be computationally challenging for large $n$. However, this may also contribute to the storage and recognition capacity of the network. In some cases, we may choose to use averaging and majority vote to decrease the number of outputs, a quite common approach (see Section \ref{sec:exp}).

The main change compared to a conventional Perceptron is that we are allowing intra-layer connections. If the output of neurons are used as inputs for other neurons, we cannot calculate local gradients, as mutual dependencies  do not allow for gradient-based local adjustments. 

\subsection{Problem 2: Neurons}
Another problem, perhaps even more severe than the previous one, in defining a complete graph as a new neural network, would be the characterization of a \emph{neuron}. Artificial neurons have weighted inputs from the input layer or other neurons from a previous layer. However, they do not have incoming weighted connections from other neurons of the same layer. As a matter of fact, some of the breakthroughs in recent years were achieved because we excluded exactly this possibility; in Boltzmann machines, by excluding connections between neurons of the same layer, \emph{restricted} Boltzmann machines were born and enabled us to train deep networks \cite{hintonRBM}. In a complete graph, however, there is no notion of a ``layer''. So what does a neuron look like in such a network?

\begin{figure}[htb]
\begin{center}
\includegraphics[width=0.99\columnwidth]{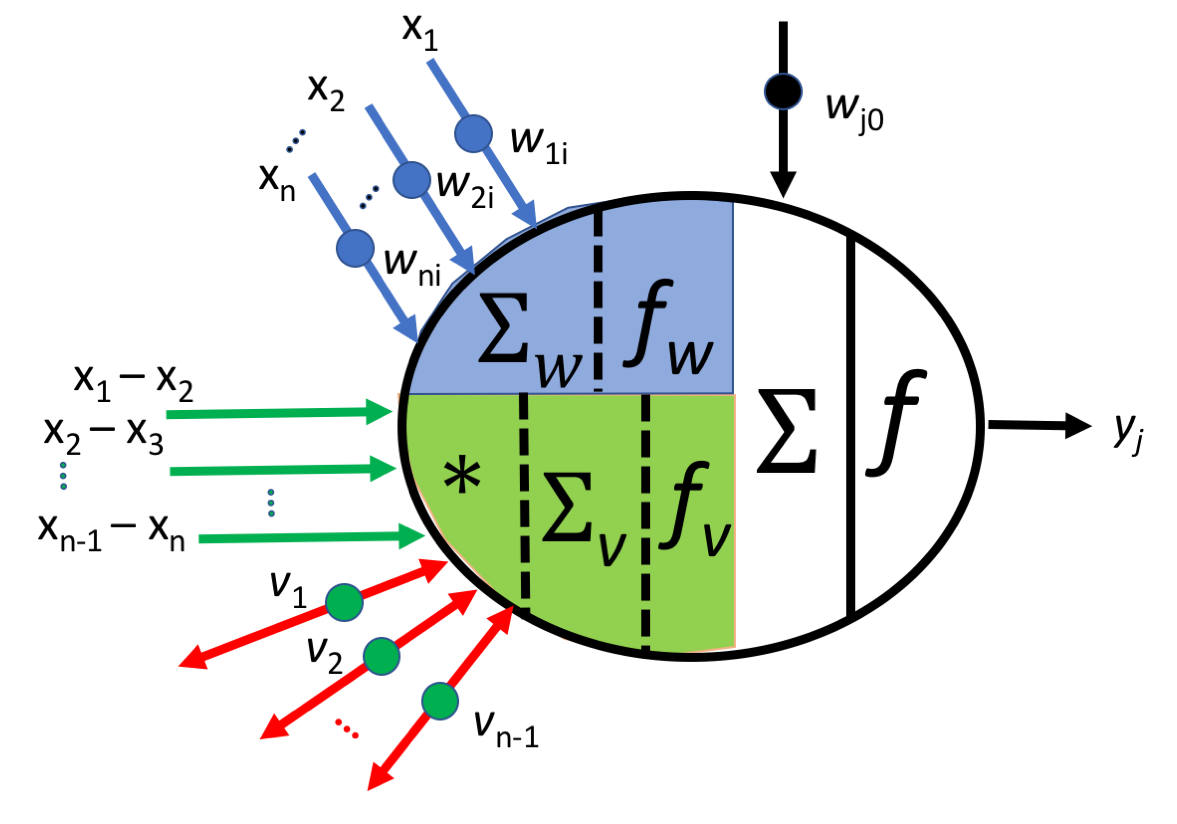}
\caption{Paraneuron -- all inputs are inserted into a processing unit as usual whereas the bidirectional connections from other neurons are also used to weight the input differences (see Equation \ref{eq:paraneuron}). The weighting of differences happens inside the unit. }
\label{fig:para}
\end{center}
\end{figure}

\begin{figure}[htb]
\begin{center}
\includegraphics[width=0.9\columnwidth]{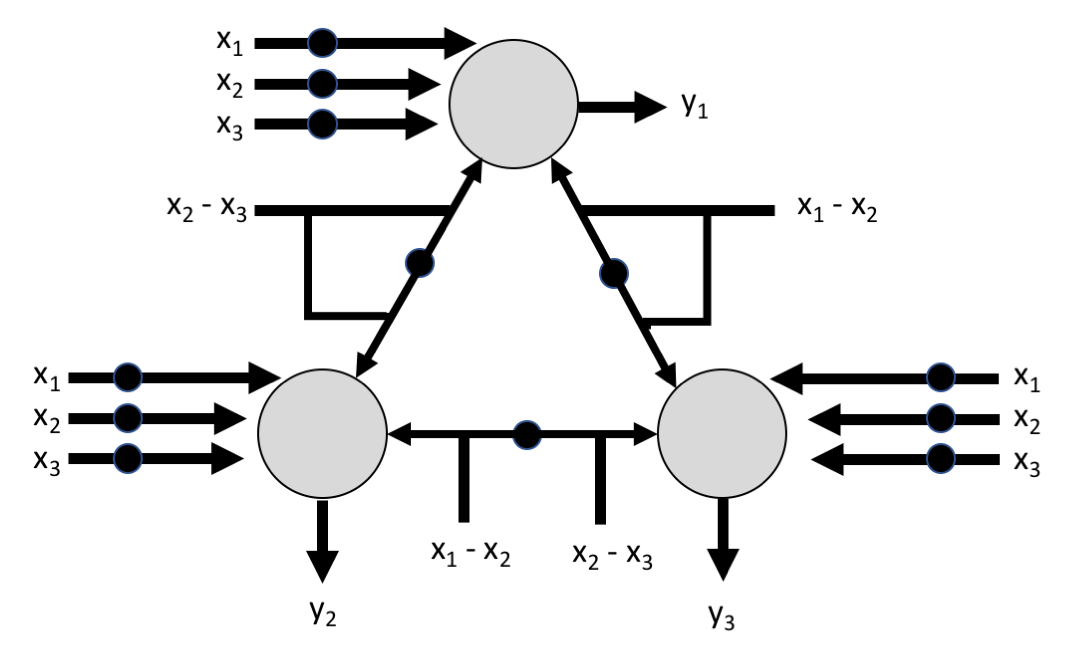}
\caption{A s-Perceptron with 3 inputs and 3 paraneurons.}
\label{fig:3n}
\end{center}
\end{figure}

Let's call, for sake of distinction, a neuron in a complete graph a \emph{paraneuron}. A paraneuron has two distinct sets of inputs: 1)  directed inputs $x_{i}$ weighted with corresponding weights $w_{ij}$ for the $i$-th input going to the $j$-th neuron, and 2)  bidirectional graph edges with shared weights $v_i$ (that make the paraneurons suitable for building a connected graph) applied on input differences $x_i-x_{i+1}$ (Fig. \ref{fig:para}). Using input differences emerged from practical design constraints: we had to assign weights to the graph edges but to give them a role rather than a collection of ``biases'' they needed to be attached to some values rather the inputs; input differences can severe as inputs as they do provide additional information about the behaviour of the inputs but their selection is empirical based on general usefulness of ``differences'' (e.g., using differences to extend genetic algorithms to \emph{differential evolution} \cite{storn1997differential}). Given three logistic functions $f(\cdot), f_w(\cdot)$, and $f_v(\cdot)$, the output $y_j$ of the $j$-th paraneuron can be given as
\begin{equation}
\label{eq:paraneuron}
y_j\!=\! f\!\left(f_v (\sum_{i=1}^{n-1} (x_i\!-\!x_{i+1})v_i )+ f_w(\sum_{i=1}^{n} x_{i} w_{ij})\!+\! w_{j0}\right),
\end{equation}
\noindent with the bias $w_{j0}$. As it can be seen in a simple example with only 3 paraneurons (Fig. \ref{fig:3n}), all inputs can be fed into all paraneurons and $n$ edges between the $n$ paraneurons are weighted and shared between connected nodes to process input differences. 

\subsection{Problem 3: Learning}
The idea of graphs and their connection to neural networks have been around for some time. The major reason for not pursuing complete graphs as the main concept of neural network implementation is perhaps the immense difficulty in training such networks. The sheer number of weights, and the existence of mutual dependencies, makes the training of a complete graph appear impossible, and not just for large $n$. To train the proposed s-Perceptron with $n$ paraneurons, we propose three key ideas: 
\begin{enumerate}
\item Adjusting the weights for large sub-graphs $G'\subset G$ (with $|V'|\approx |V|$) at the beginning of learning and continue the learning by progressively making the sub-graphs smaller and smaller until $|V'|\ll |V|$, 
\item the sub-graphs $G'$ are selected randomly, and 
\item the adjustments are \emph{graph-wise}, meaning that we chose between the sub-graph $G'$ and its modified version $\hat{G}'$ to either update the graph with the weights of $G'$ (hence, no change) or with the weights of its modified version $\hat{G}'$ (hence, adjustment) depending on the total error of the graph $G$. 
\end{enumerate}

The network error is as always the driving force behind the learning. The main design challenge, however, is to define the modification that generates $\hat{G}'$.

\section{Experiments}
\label{sec:exp}
\begin{figure*}[htb]
\begin{center}
\includegraphics[width=0.75\textwidth]{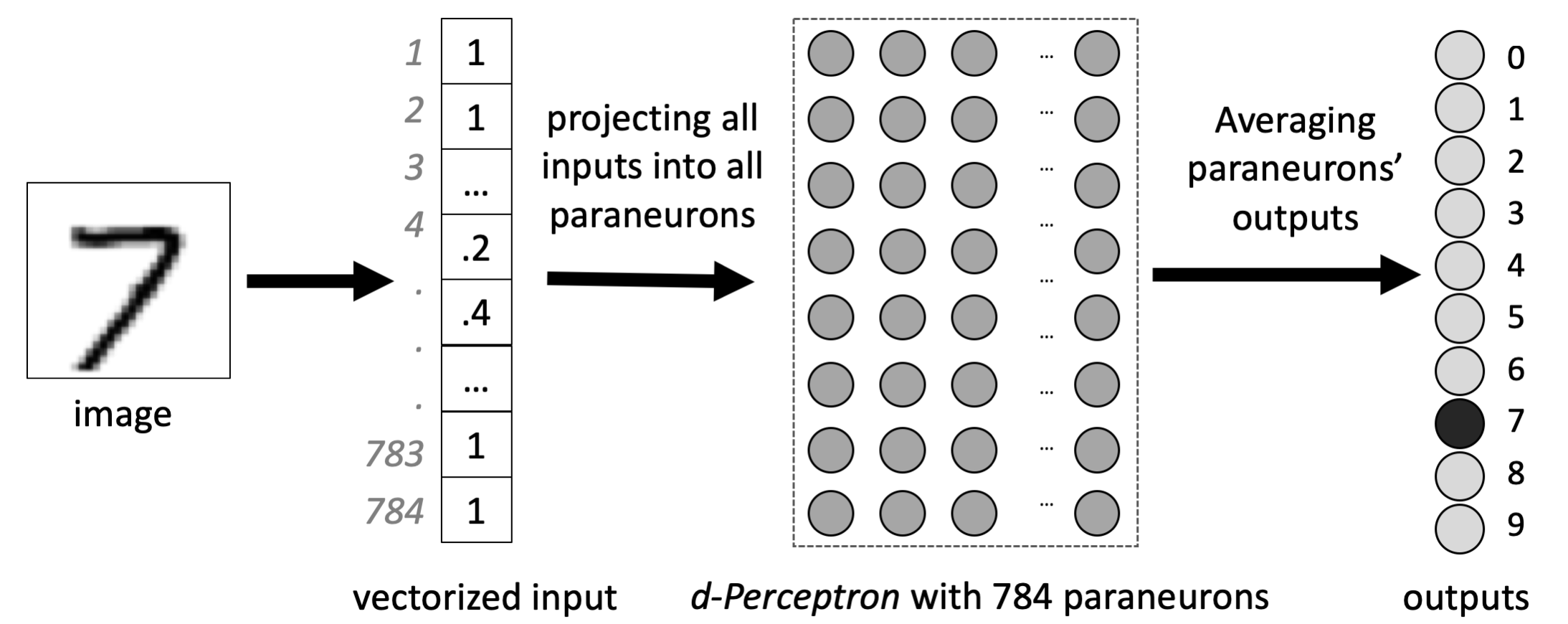}
\caption{Using s-Perceptron for digit recognition: The setup for using a s-Perceptron for recognizing digits from the MNIST dataset.}
\label{fig:digits}
\end{center}
\end{figure*}

To conduct preliminary tests on the proposed s-Perceptron, we selected ``digit recognition''. Handwritten digit recognition is a well-established application for neural networks.   The configuration for digit recognition with the  s-Perceptron is illustrated in Fig. \ref{fig:digits}. The main change for using the s-Perceptron for digit recognition is that we average $\lfloor \frac{n}{10}\rfloor$ outputs to generate probabilities for digits. Fully connected layers have a large number of parameters and are prone to overfitting. In the architecture of s-Perceptron we avoid using the fully connected at the end of the network. Instead, we are using the global average pooling. Global average pooling groups adjacent neurons into multiple groups and calculate their average values. The average values from all the groups are activated with Softmax function to calculate the output class distribution. The average operation itself doesn't have any parameter thus tend not to overfit as much as fully connected layers. In fact, many modern CNN architectures these days tend to use global average pooling instead of fully connected layers. For examples, DenseNet, SqueezeNet, Inception Network.

\subsection{MNIST Dataset}
\begin{figure}[h]
\begin{center}
\includegraphics[width=0.9\columnwidth]{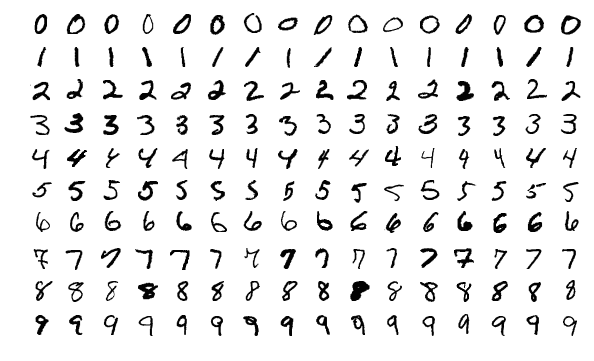}
\caption{Examples from MNIST dataset.}
\label{fig:mnistexamples}
\end{center}
\end{figure}

The MNIST dataset is one of the most popular image processing datasets and contains several thousands  handwritten digits \cite{lecun1998mnist} (Fig. \ref{fig:mnistexamples}). In particular, there are a total of 70,000 images of size $28\times28$ depicting digits 0 to 9. The dataset has a preset configuration of 60,000 images for training and 10,000 images for testing. The s-Perceptron, hence, will have $n=28\times 28=784$ paraneurons.

\subsection{Training via Gradient Descent}
The proposed architecture is a graph. However, since no paraneuron's output is used as input to another paraneuron, the gradient descent algorithm is still applicable. The proposed s-Perceptron was first implemented using the  TensorFlow~\cite{tensorflow2015-whitepaper} library which supports automatic differentiation of computational graphs. Paraneurons were implemented according to~Fig.~\ref{fig:para} with a slight  modification. The modification involves -- the addition of an extra term to the input differences, i.e. $x_n - x_1$ for the simplification of the code. The addition of an extra term balances the length of input $x$ and input differences $x_n - x_{n+1}$. With the preceding modification, equation ~\eqref{eq:paraneuron} can be rewritten in form of matrix multiplication as $y_j = f\left(f_v \left(x_d v \right)+ f_w\left(xw\right)\right)$,
where $x_d$ is input differences, $v$ \& $w$ are weight matrices of size $|x| \times |x|$, and $v$ is a symmetrical matrix. The symmetric constraint of $v$ can be compensated by substituting $v$ with  $\frac{v + v'}{2}$. Rectified Linear Unit (ReLU) is chosen as the activation function $f_v$ and $f_w$. For positive inputs, this function will return the input itself, otherwise it will return zero. Since the output of a ReLU function is a positive number, the activation function $f$ can be ignored.

\begin{figure}[ht]
\begin{center}
\includegraphics[width=0.9\columnwidth]{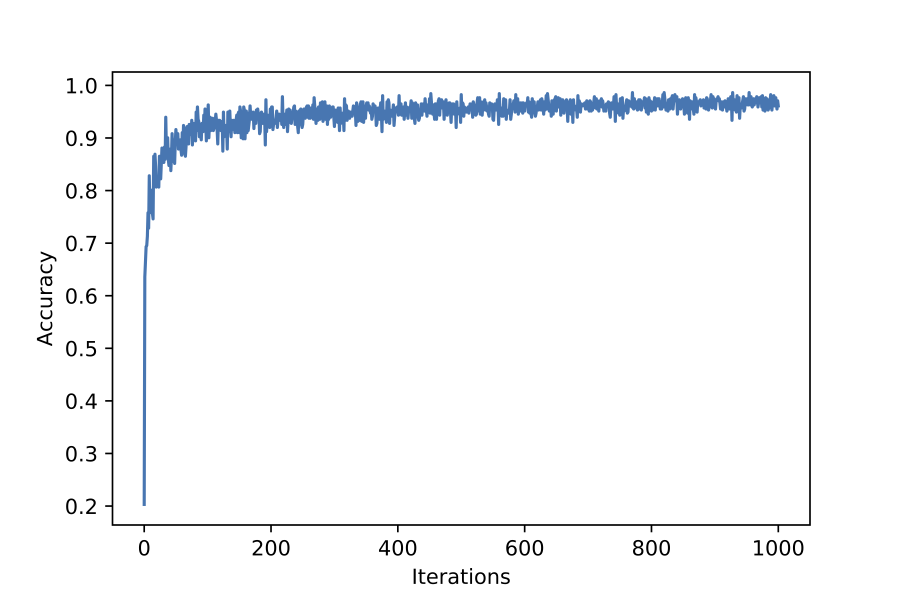}
\caption{The training accuracy of s-Perceptron for gradient descent algorithm.}
\label{fig:tf_acc}
\end{center}
\end{figure}
 Fig.~\ref{fig:tf_acc} shows the accuracy of the model for each iteration during the training phase. The maximum training accuracy achieved by the model is $98.63\%$. Whereas the accuracy reported in literature for a 2 layer MLP on MNIST dataset is $98.4\%$~\cite{simard2003best}. The higher accuracy of the s-Perceptron justifies its learning capacity. The number of trainable parameters in the model is $784 \times 784 + \frac{784 \times 784}{2} = 921984$. There is a huge literature on MNIST dataset with a leader-board reporting all major results \footnote{http://yann.lecun.com/exdb/mnist/}. Ciresan et al. reported 99.77\% accuracy using en ensemble of $n$ multi-column deep convolutional neural networks followed by averaging \cite{cirecsan2012multi}. Wan et al. used DropConnect on a network with $700\times 200\times 10=140,000,000$ weights and reported highest accuracy equal to 99.79\% \cite{wan2013regularization}. However, when simpler structures are used, accuracy cannot exceed 98.88\% (Table 2 in \cite{wan2013regularization}). Such numbers were achieved with two layer-networks with $800\times 800\times 10=6,400,000$ weights. This a network 14 times larger than the s-Perceptron.

\textbf{Critical Summary --}
The extensions of the historic Perceptron may have the potential to become the building block of a new generation of artificial neural networks. The s-Perceptron, however, suffers from multiple shortcomings that should be addressed in future works:
\begin{itemize}
    \item The s-Perceptron becomes a realistic graph implementation when the axons of paraneurons become the dendrites to other paraneurons. In its current form, the s-Perceptron is just a \emph{graph-like} topology. 
    \item Vectorizing the image to be used as the input for the s-Perceptron limits the applications to cases where small images are used or downsampling is not a disadvantage toward accuracy. Employing features, or some type of embedding, instead of vectorized raw pixels could be more effective. This may be regarded as the imitation of the optic nerve transferring the encoded retinal information to the visual cortex.  
    \item Using input differences as inputs for paraneurons seems to be a simple (and arbitrary) remedy to extend Perceptrons to graphs. However, investigations need to assess whether processing the differences is necessary or not.
    \item The output of the s-Perceptron needs to be customized for each application. This is due to the assumption that all paraneurons do contribute to the output.  
    \item Although we did not test the case when a neurons/paraneurons exhibit dependencies,it seems gradient descent would not work anymore. Hence, investigating non-gradient based learning is necessary. 
    \item Graphs do inherently provide more \emph{access points} for being integrated within a larger network. However, this does not mean that embedding subgraphs (for small tasks) inside larger graphs (for a multitude of tasks) is straightforward. The research in this regard is still in its infancy.  
\end{itemize}
Beyond gradient descent, we also experimented with new learning algorithms based on opposition \cite{al2010opposition} where the weights were adjusted by switched between their values and \emph{opposite values} simultaneously shrinking the search interval. However, we could not achieve accuracy values higher than 50\%. 

\section{Conclusions}
In this paper, we introduced the notion of a  \emph{subtractive Perceptron} (short s-Perceptron), a graph implementation of a  Perecptron which in addition to existing inputs  also uses the pairwise input differences as shared edge weights between its neurons, called paraneurons. The s-Perceptron is an undirected graph with no paraneuron's input coming from another paraneuron, to investigate a simple graph topology. The motivation for working on graph-based networks is to create a more realistic imitation of the human brain's architecture; a capable d-Perceptron could be the building block for a much larger network (i.e., brain) that learns one specific task within that larger network. The proposed s-Perceptron delivers good results if trained with the gradient descent algorithm. However, if we extend it to a directed graph, then we may have to abandon gradient descent and find other learning algorithms to solve the credit assignment problem. 


\bibliographystyle{ieee}
\bibliography{egbib}

\end{document}